\documentclass[letterpaper]{article} 
\usepackage{aaai2026}  
\usepackage{times}  
\usepackage{helvet}  
\usepackage{courier}  
\usepackage[hyphens]{url}  
\usepackage{graphicx} 
\usepackage{natbib}  
\usepackage{caption} 
\usepackage{algorithm}
\usepackage[utf8]{inputenc} 
\usepackage[T1]{fontenc}    
\usepackage{url}            
\usepackage{booktabs}       
\usepackage{amsfonts}       
\usepackage{nicefrac}       
\usepackage{microtype}      
\usepackage{xcolor}         
\usepackage{bm}
\usepackage{pifont}
\usepackage{color}
\usepackage{multirow}  
\usepackage{xspace}
\usepackage{amsmath}
\usepackage{amssymb}
\usepackage{float}
\usepackage{algorithm}
\usepackage{algorithmicx}
\usepackage{algpseudocode}
\usepackage{amsmath,amssymb} 
\usepackage{subfloat}
\usepackage{tabularx}
\newenvironment{icompact}{
  \begin{list}{$\bullet$}{
    \itemindent -.05em
    \parsep 0pt plus 1pt
    \partopsep 0pt plus 1pt
    \topsep 2pt plus 2pt minus 2pt
    \itemsep 0pt plus 1.3pt
    \parskip 0pt plus 2pt
    \leftmargin 0.13in}
      }
{\normalsize\end{list}}
\newcommand{\sys}{\textsc{FedExchange}\xspace}

\usepackage{caption}
\usepackage{graphicx}
\usepackage{svg}
\usepackage{soul}
\usepackage{subfig}
\usepackage{diagbox}

\title{}

\usepackage{newfloat}
\usepackage{listings}
\DeclareCaptionStyle{ruled}{labelfont=normalfont,labelsep=colon,strut=off} 
\floatstyle{ruled}
\newfloat{listing}{tb}{lst}{}
\floatname{listing}{Listing}
\title{\sys: Bridging the Domain Gap in Federated Object Detection for Free}
\author {
    Haolin Yuan\textsuperscript{\rm 1},
    Jingtao Li\textsuperscript{\rm 2},
    Weiming Zhuang\textsuperscript{\rm 2}
    Chen Chen\textsuperscript{\rm 2}
    Lingjuan Lyu\textsuperscript{\rm 2}
}
\affiliations {
    \textsuperscript{\rm 1} Johns Hopkins University\\
    \textsuperscript{\rm 2} Sony AI\\
hyuan4@jhu.edu, \{jingtao.li, weiming.zhuang, chen.chen, lingjuan.lyu\}@sony.com
}

\begin{document}

\maketitle
\begin{abstract}
Federated Object Detection (FOD) enables clients to collaboratively train a global object detection model 
without accessing their local data from diverse domains. However, significant variations in environment, weather, and other domain-specific factors hinder performance, making cross-domain generalization a key challenge.
Existing FOD methods often overlook the hardware constraints of edge devices and introduce local training regularizations that incur high computational costs, limiting real-world applicability.
In this paper, we propose \sys, a novel FOD framework that bridges domain gaps without introducing additional local computational overhead. \sys employs a server-side dynamic model exchange strategy that enables 
each client to gain insights from other clients' domain data without direct data sharing. Specifically, \sys allows the server to alternate between model aggregation and model exchange. During aggregation rounds, the server aggregates all local models as usual. In exchange rounds, \sys clusters and exchanges local models based on distance measures, allowing local models to learn from a variety of domains. As all operations are performed on the server side, clients can achieve improved cross-domain utility without any additional computational overhead.
Extensive evaluations demonstrate that \sys enhances FOD performance, achieving 1.6$\times$ better mean average precision in challenging domains, such as rainy conditions, while requiring only 0.8$\times$ the computational resources compared to baseline methods. 
\end{abstract}
\section{Introduction}
\label{sec:Introduction}

Federated Learning (FL)~(\citealt{FedAvg, FL_classification_1, FL_2, FL_classification_2, FL_classification_3, FedSGD}) is a privacy-aware learning framework that facilitates collaborative training among clients while maintaining data locality and privacy. 
Over recent years, FL has shown great potential in diverse fields, including healthcare for predictive diagnostics~(\citealt{FL_classification_3, health_1, health_2}), finance for fraud detection~(\citealt{finance_1, finance_2}), and various computer vision applications~(\citealt{COALA, cross-domain_fL}).


Although FL has been widely adopted in various computer vision tasks, its application to object detection~\citep{cross_domain_OD_1, cross_domain_OD_4} presents distinct challenges and remains largely under-explored. Object detection is an important task for numerous high-impact applications, including autonomous driving~(\citealt{auto}), surveillance systems~(\citealt{surveillance, yang2024anomalyruler}), augmented reality~(\citealt{AR, AR_2}), and robotics~(\citealt{ob_robot_1,ob_robot_2}). Unlike classification or segmentation, object detection demands precise localization and recognition of multiple objects, often at varying scales and spatial configurations. These requirements introduce substantial algorithmic and computational complexity. In addition, the annotation process for object detection is highly labor-intensive, as it involves detailed bounding box annotations for each object instance. This overhead often results in limited labeled data, which in turn hinders the generalization ability of models across diverse domains and real-world scenarios.

Given these challenges, federated learning (FL) presents a promising paradigm by enabling collaborative model training across distributed data silos without sharing raw data. However, the presence of significant domain heterogeneity among clients makes its application to object detection particularly complex. In contrast to classification tasks—where heterogeneity often arises from differences in class distribution—object detection encounters more profound discrepancies in visual characteristics. For example, data captured by autonomous vehicles or surveillance systems can vary dramatically in viewpoints, geographic locations, and environmental conditions~\citep{cross_domain_OD_1, cross_domain_OD_2, cross_domain_OD_3, cross_domain_OD_4}. Such domain-specific variations pose substantial challenges for conventional FL algorithms, highlighting the necessity of designing FL frameworks specifically tailored to the demands of object detection.

\begin{figure}
  \centering
  \includegraphics[width=0.48\textwidth]{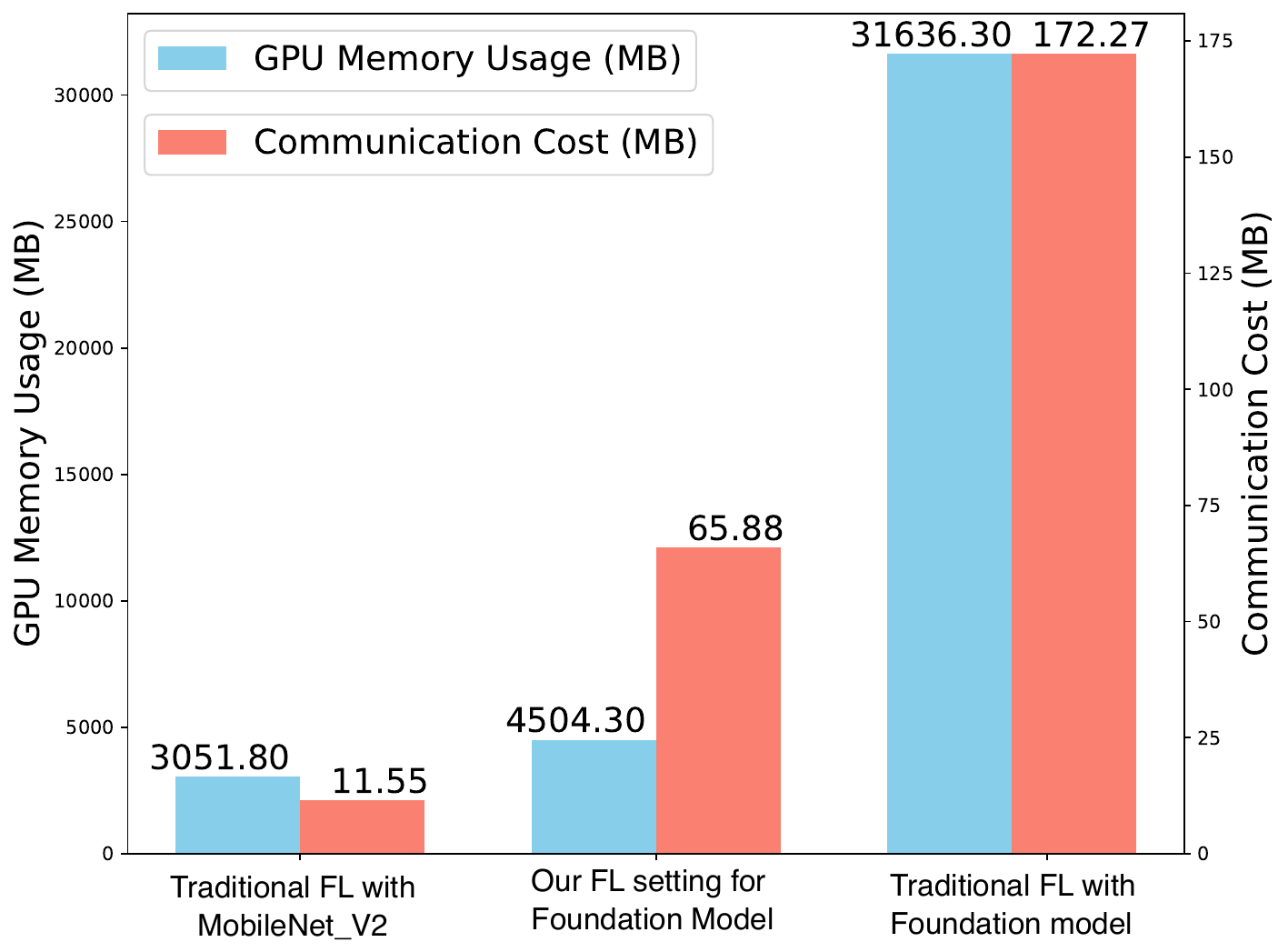}
  \caption{Comparison of memory and communication costs between (1) traditional FL settings using foundation models with fully learnable backbones and decoders, (2) lightweight models such as MobileNet, and (3) our practical federated object detection (FOD) setting with a frozen backbone and learnable decoder. Notably, a fully learnable ViT-adapter plus decoder setup incurs approximately 7× the GPU memory usage and 3× the communication cost compared to our proposed setting.}
  \label{fig:motivation}
  \vspace{-0.2in}
\end{figure}

Existing approaches~\citep{FedSTO} primarily focus on training a generalized model that performs reasonably well across diverse client domains. However, due to the inherent variability in cross-domain data and the nature of weighted model aggregation, clients with limited or significantly different data distributions are often marginalized. As demonstrated in our evaluation (Section~\ref{sec:experiment}), after the global aggregation, a client in a cross-domain person detection task achieves only 13.58 mean average precision (mAP), compared to the global mAP of 30.57, revealing considerable disparities in model performance. 

Moreover, current work~\citep{FedSTO} overlooks the practical constraints imposed by real-world federated object detection settings. Edge devices, which are commonly used in such deployments, often lack the capacity to train even lightweight object detection models. Although models like MobileNet-V2~\citep{mobilenet_v2} paired with SSDlite~\citep{SSD} are designed for efficiency, devices such as the Nvidia Jetson Nano~\citep{jetson_nano} can barely support training them. Despite the small model footprint, Figure~\ref{fig:motivation} shows that GPU memory consumption becomes prohibitive when training on batches of 16 with 480$\times$640 resolution images, largely due to the computational demands of backpropagation. These limitations underscore the importance of developing FL systems that are both domain-adaptive and computationally efficient.

With the emergence of foundation models (FMs) in vision~\citep{zhuang2023foundation}, which exhibit strong adaptability across diverse visual domains and tasks, we propose a practical setting that addresses key real-world constraints. In our framework, a foundation model backbone is first distributed to all clients. Each client then initializes a local object detection decoder for training and participates in federated aggregation. As illustrated in Figure~\ref{fig:motivation}, freezing the FM backbone introduces minimal GPU memory overhead, resulting in consumption comparable to that of MobileNet-V2, while retaining the advanced feature extraction capabilities of the FM. This setup offers two key benefits: it significantly reduces computational costs and lowers communication overhead, making the framework particularly suitable for deployment in edge-based federated object detection scenarios.

\begin{figure*}[!t]
\centering
{\includegraphics[width= 0.9\linewidth]{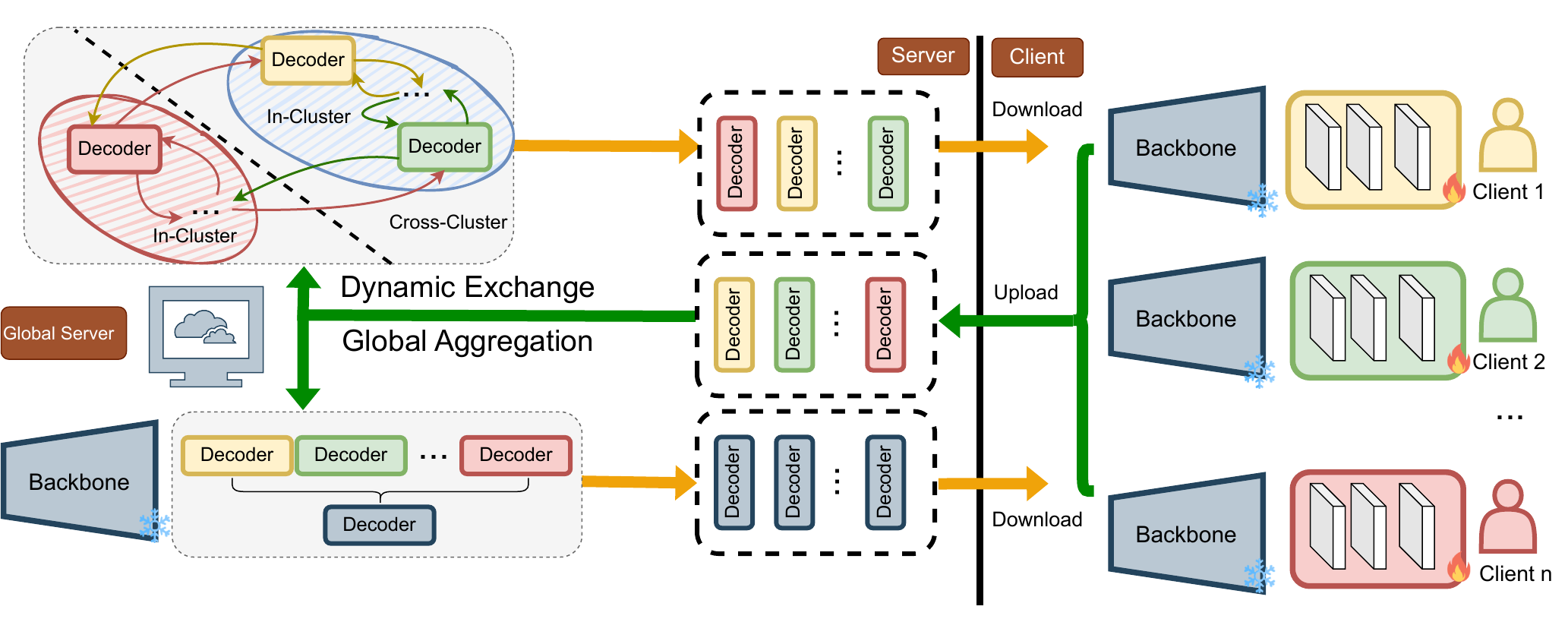}}
\caption{Workflow of \sys. The global server first defines the aggregation frequency $T$. During FOD training, clients upload their local model weights to the server at each communication round. Based on the current round, the server determines whether to perform model aggregation or dynamic model exchange. If dynamic exchange is selected, the server clusters all uploaded models into two groups using cosine distance and performs both in-cluster and cross-cluster exchanges. The server then sends the resulting model—either an exchanged or aggregated version—back to each client.}  
\label{fig:procedure}
\vspace{-0.1in}
\end{figure*}

Additionally, the decoder—typically composed of a Region Proposal Network (RPN) head and a Region of Interest (RoI) head—provides more effective feature extraction than the linear layers commonly used in classification tasks. This motivates the development of a practical and efficient solution tailored to federated object detection (FOD). We introduce \sys, a novel FOD framework designed to improve model performance across heterogeneous domains without introducing additional computational overhead on clients. The key innovation of \sys is a dynamic model exchange mechanism that enables models to acquire cross-domain knowledge through a server-side, cluster-based exchange strategy. After clients upload their decoder weights, the global server decides whether to perform aggregation or model exchange. For aggregation, \sys follows the standard FedAvg protocol~\citep{FedAvg}. For model exchange, the server clusters decoders based on a distance metric and performs both intra-cluster and inter-cluster exchanges, as illustrated in Figure~\ref{fig:procedure}. To ensure diversity, clients receive decoders different from those in the previous round. This approach encourages cross-domain learning without compromising data privacy.
By exchanging decoders trained on different domains, \sys facilitates domain-invariant feature learning and supports the construction of a more robust and generalized global model.

Our contributions can be summarized as follows:
\begin{itemize}
    \item To the best of our knowledge, \sys is the first practical framework for addressing cross-domain challenges in Federated Object Detection (FOD) without incurring additional local computational overhead.
    \item \sys introduces a dynamic model exchange mechanism that leverages server-side clustering based on distance metrics, enabling in-cluster and cross-cluster decoder exchanges to facilitate cross-domain knowledge transfer.
    \item Extensive experiments on six cross-domain object detection benchmarks demonstrate that \sys significantly improves performance—achieving up to 1.6$\times$ higher mean average precision in challenging domains such as rainy conditions—while using only 0.8$\times$ the GPU resources of baseline methods.
\end{itemize}


\section{Related Work}

\paragraph{Federated Learning (FL).} Federated learning (FL)~(\citealt{FL_classification_1, FL_classification_2, FL_classification_3, FedAvg}) has been widely adopted to train generalized models across diverse domain datasets. However, achieving strong performance in FL remains challenging in the presence of heterogeneous local data distributions. To address this, numerous studies have explored cross-domain solutions within FL. For example, \citealt{cross-domain_fL} propose personalized FL approaches that allow clients to maintain tailored models, while \citealt{cross-domain_fL_2} introduce a variance-aware dual-level prototype clustering method to reduce representation variance across domains. Similarly, \citealt{cross-domain_fL_3} present a discriminative transfer technique based on feature and label consistency for domain adaptation. Compared to classification tasks, applying FL to object detection, i.e., Federated Object Detection (FOD), 
introduces greater complexity and receives limited attention, largely due to the inherent difficulty of handling diverse domain distributions and data scarcity as it requires precise spatial annotations and the higher variability in visual data. 
Recently, Kim et al. proposed FedSTO~\citep{FedSTO}, a semi-supervised FOD framework that addresses cross-domain challenges. 

\paragraph{Cross-domain Object Detection.}
Cross-domain object detection aims to reduce performance gaps between source and target domains. Existing approaches can be broadly categorized into two main paradigms: feature alignment and knowledge distillation. Feature alignment methods focus on increasing the similarity of features extracted from different domains, enabling models to learn domain-invariant representations that generalize effectively. For example, \citealt{strong_weak_alignment} propose a strong-weak alignment strategy, while \citealt{center_alignment} leverage center alignment, and \citealt{multi-level_alignment} employ multi-level feature alignment. Other works ~(\citealt{other_feature_alignment_1, other_feature_alignment_2, other_feature_alignment_3, other_feature_alignment_4}) incorporate additional constraints during adversarial training to improve cross-domain feature alignment. In contrast, knowledge distillation methods train a student model to mimic the behavior of a teacher model, often one trained on the target domain. For instance, \citealt{unbiased_KD} address model bias using soft labels and instance selection in a mean-teacher framework, while \citealt{dual_branch_KD} propose a dual-branch structure that supports mutual learning between teacher and student via a shared classifier. Despite their effectiveness, these methods are not directly applicable in federated object detection (FOD) settings, as they require access to target domain samples—violating the fundamental privacy constraints in federated learning.

\section{Method}


\subsection{Problem Formulation and Overview}

\paragraph{Problem Formulation.} \sys views each foudation model as comprising two distinct components: a model backbone $f(\cdot)$, which is pre-trained on public datasets to develop feature extraction capabilities and kept frozen during the entire FOD training, and a learnable decoder $g(\cdot)$, which is placed after the backbone to generate final predictions. Formally, given any input image $x$ and its corresponding ground truth label $y$, the overall model $W$ can be represented as:
\begin{equation}
  W(x,y) = g(f(x,y))
\end{equation}
where $f(x)$ extracts features from $x$, and $g(f(x))$ applies the decoder to these features to generate a proposal list and produce the final prediction. 

The goal of the FOD training is to collaboratively train a universal decoder that is generalized on clients' data from different domains. Specifically, the optimization goal is to minimize
\begin{equation}
    min_g \mathcal{L}(g) = \sum^m_{i=1} \frac{n_i}{n}\mathcal{L}_i(g)
\end{equation}
 where $n_i = |D_i|$ is the size of local dataset for client $i$, $n = \sum^m_{n=1}n_i$, and $\mathcal{L}_i(g) = \mathbb{E}_{(x,y)\sim D_i}[\ell_i(x,y;g)]$.

\paragraph{Overview.} The core concept underlying \sys is motivated by the empirical observation that centralized learning typically yields superior performance compared to federated learning in scenarios characterized by cross-domain data distributions.  Accordingly, \sys is designed to emulate multiple centralized learning processes concurrently without the sharing of private local data among clients. \sys framework comprises two principal stages: global-end clustering (section~\ref{sec:global-clustering}) and dynamic model exchange (section~\ref{sec:model-exchange}). 
Initially, in the global-end clustering phase, the central server partitions all uploaded decoders into distinct clusters based on a pre-defined distance metric between model pairs. Subsequently, the dynamic model exchange stage executes the exchange of models both within and across clusters, ensuring that clients are assigned models different from those they received in the previous round.

\subsection{Global-End Clustering}
\label{sec:global-clustering}

\sys enables the global server to pre-define the aggregation frequency $T$, which determines whether the server performs clustering and exchange on the uploaded decoders or aggregates them. 
Given a total of R training rounds, \sys mandates that $R\%T=0$ to ensure clients receive the aggregated global decoder in the final round. Specifically, at a given round $r$  and frequency $T$, if $r\%T=0$, the global server executes a global aggregation; otherwise, it performs an exchange operation on all decoders.

Initially, \sys treats each decoder as an individual cluster, subsequently merging clusters iteratively until only two clusters remain.
This default setting of two clusters is adopted due to the inherent difficulty in predefining the optimal number of clusters for diverse participating data domains. Should certain models be better separated under an optimal clustering than grouped together, the in-cluster exchange mechanism detailed in Section~\ref{sec:model-exchange} is designed to handle them appropriately.
Specifically, for all decoders $G=\{g_1,g_2,g_3,...,n_n\}$ , \sys first calculates the cosine distance $d_C(g_i,g_j)$ between every decoder pairs $g_i, g_j$, $i,j \in \{1,2,3,...,n\}$, 
\begin{equation}
    d_C(g_i, g_j) = 1- \frac{g_i \cdot g_j}{||g_i||||g_j||}   
\label{eq:1}
\end{equation}
where $||g_i||=\sqrt{\sum_{n=0}{g_n}^2}$ is the magnitude of $g_i$. \sys then forms an $N \times N$ distance matrix based on all $d_C(g_i,g_j)$. During the iterative merging process, \sys looks for two clusters that have the smallest average linkage distance, which is defined as 
\begin{equation}
    d_{AL}(C_i, C_j) = \frac{1}{|C_i||C_j|}\sum_{g_u\in C_i}\sum_{g_v\in Cj}d_C(g_u,g_v)
\label{eq:2}
\end{equation}
\sys merges these clusters to form a new cluster, and updates the distance matrix accordingly. It then repeats computing Equation~\ref{eq:1} and Equation~\ref{eq:2} till the final two clusters are formed.

\begin{figure}[!t]
\centering
{\includegraphics[width= 0.8\linewidth]{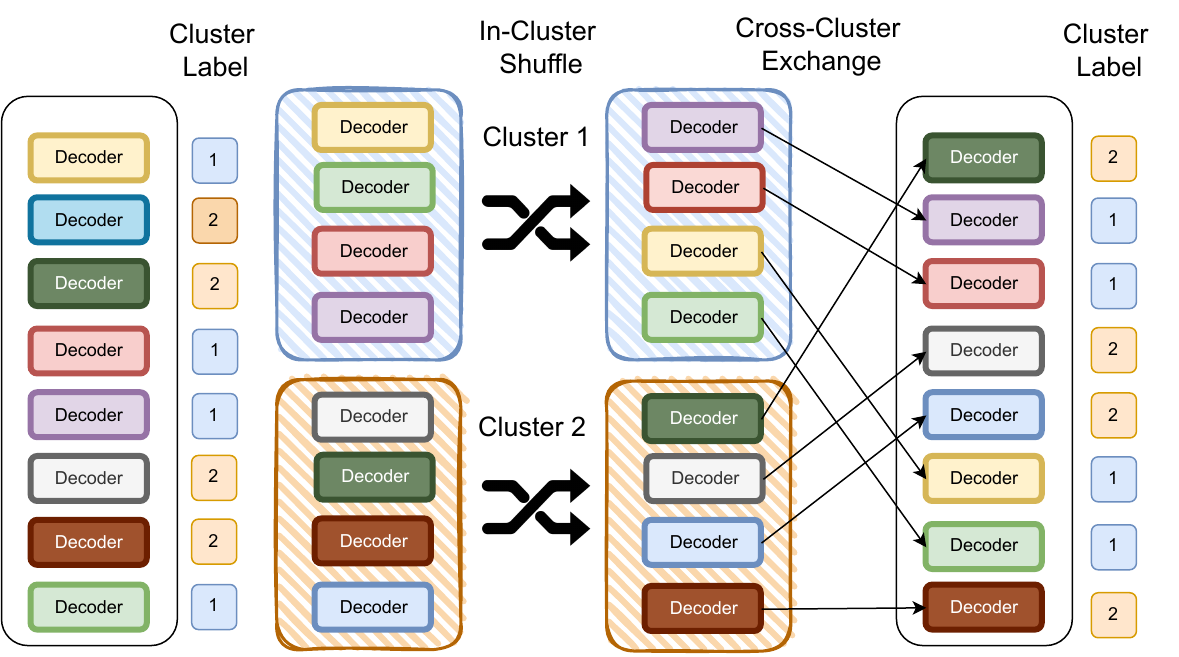}}
\caption{An illustration of dynamic model exchange workflow. \sys first splits all models into two clusters based on pair-wise cosine distance. \sys shuffles models in each cluster to achieve in-cluster model exchange. Then for cross-cluster model exchange, \sys goes through the original model list, checks the cluster label of each decoder, and assigns a decoder from the other cluster to the same client. }  
\label{fig:dynamic_exchange}
\end{figure}

\subsection{Dynamic Model Exchange}
\label{sec:model-exchange}

We show the procedure of dynamic model exchange in Figure~\ref{fig:dynamic_exchange}. Following the clustering process, \sys generates a cluster index list, $I$, corresponding to the original list of decoders. For any decoder $g_i \in G$,  , the cluster index $I[i]$ is defined as:

\begin{equation}
    I[i] = 
    \begin{cases} 
    0 & \text{if } g_i \in C_0 \\
    1 & \text{if } g_i \in C_1 \\ 
    \end{cases}
\end{equation}
Decoders within the same cluster are considered to originate from similar data domains, whereas decoders across different clusters are considered to represent distinct data domains. 
To facilitate the acquisition of insights from similar domains among models in the same clusters, \sys employs a random shuffling of models within each cluster, thereby achieving in-cluster model exchange.  
For cross-cluster exchange, \sys reverses each element in the cluster index list $I$ by applying the transformation $1-I[i]$. This operation ensures that decoders from cluster $C_0$ are assigned to clients who received decoders from cluster $C_1$ in the preceding round. 

Two scenarios worth consideration following the clustering. If $|C_0|=|C_1|$, the models are divided into clusters of equal size. In this case, during the dynamic exchange, each decoder $g_i \in C_0$ is paired with a model from $C_1$, resulting in exclusively cross-cluster exchanges. Conversely, if $|C_0| \neq |C_1|$, there will inevitably be decoders that are not paired with a cross-cluster counterpart. The preceding random shuffle within clusters ensures that such decoders can be assigned to clients within their original cluster.


We provide pseudo-code of \sys in Algorithm~\ref{algorithm} in Appendix. Prior to the first training round, the global server pre-trains the backbone model on public datasets and distributes this frozen backbone, along with the initialized decoder, to all participating clients.
During the local training, i). All clients train their decoders on their respective local domains (line 7); ii). Clients upload their updated decoder weights upon completing local training (line 8); iii). The global server receives all local models; iv). Based on the current round number and the pre-defined aggregation frequency $T$, the server either performs decoder exchange or aggregation to produce the global decoder (line 10-line 14); v).The global server then sends the resulting decoders back to the clients (line 17-line 20), and this process repeats from step i) until the specified number of training rounds is completed.

\section{Experiments}
\label{sec:experiment}

In this section, we first present the performance of \sys on benchmark datasets compared to baseline methods in Section~\ref{sec:performance}. Section~\ref{sec:convergence} and Section~\ref{sec:GPU} discuss the convergence behavior, GPU memory usage, and communication cost of \sys, respectively. Section~\ref{sec:scare_training} examines its effectiveness under limited training data, and Section~\ref{sec:ablation} provides an ablation study. For brevity and clarity, the experimental setup and dataset details are provided in Appendix~\ref{sec:appendix}.


\begin{table*}[!h]
    \centering
    \setlength{\tabcolsep}{5pt}
    \caption{Performance comparison of \sys with baselines on cross-domain person datasets. Each client uses a unique dataset. We report the results in mAP. Best results are in bold and the second best are underscored. Our proposed \sys generally achieves better performance than the other methods.}
    \resizebox{\linewidth}{!}{
    \begin{tabular}{c|cccc|cc}
    \toprule  
     & CityPerson & WiderPerson &  CrowdHuman & COCOPerson & Average & Standard Deviation\cr
    \midrule
    FedAvg & 13.58 & \textbf{36.36} & 31.78 & 40.57 & 30.57 & 10.29\cr
    FedProx & 17.63 & 34.03 & 30.87 & 39.58 & 30.53 & 8.07\cr
    FedSTO & \underline{18.38} & 35.88 & \underline{31.83} & \underline{41.34} & \underline{31.85} & 9.79\cr
    \midrule
    \sys & \textbf{18.70} & \underline{36.07} & \textbf{32.17} & \textbf{41.38} & \textbf{32.08} & 8.39\cr
    \bottomrule
    \end{tabular}
    }
    \label{tab:person}
\end{table*}

\begin{table*}[!h]
    \centering
    \setlength{\tabcolsep}{5pt}
    \caption{Performance comparison of \sys with baselines on BDD100K. Each client contains data representing one scene of the city. \sys generally achieves superior performance than the other methods.}
    \resizebox{0.9\linewidth}{!}{
    \begin{tabular}{c|cccc|cc}
    \toprule  
     & City Street &  Highway & Residential & Mixed  & Average & Standard Deviation \cr
    \midrule
    FedAvg & \underline{28.34} & \underline{25.18} & \textbf{28.58} & 26.65 & \underline{27.19} & 1.23 \cr
    FedProx & 27.27 & 23.83 & 27.89 & 25.03 & 26.01 & 1.90 \cr
    FedSTO & 27.58 & 24.37 & \underline{28.00} & \underline{26.99} & 26.74 & 1.41 \cr
    \midrule
    \sys & \textbf{28.75} & \textbf{26.83} & 25.85 & \textbf{29.14} & \textbf{27.64} & 1.21 \cr
    \bottomrule
    \end{tabular}
    }
    \label{tab:bdd}
\end{table*}

\begin{table*}[!h]
    \centering
    \setlength{\tabcolsep}{5pt}
    \caption{Performance comparison of \sys with baselines on SODA10M. Each client contains data representing one weather condition. \sys outperforms the other methods on all domains.}
    \resizebox{0.7\linewidth}{!}{
    \begin{tabular}{c|ccc|cc}
    \toprule  
     & Clear & Overcast &  Rainy  & Average & Standard Deviation\cr
    \midrule
    FedAvg & 29.31 & 34.15 & 47.02 & 37.08 & 7.40 \cr
    FedProx &  \underline{29.86} & \underline{35.14} & \underline{47.90} & \underline{37.63} & 6.50 \cr
    FedSTO & 29.50 & 34.81 & 46.85 & 37.29 & 6.30 \cr
    \midrule
    \sys & \textbf{30.86} & \textbf{38.16} & \textbf{75.08} & \textbf{40.10} & 19.35\cr
    \bottomrule
    \end{tabular}
    }
    \label{tab:soda10m}
\end{table*}

\subsection{Performance Comparison}
\label{sec:performance}
In this section, we compare \sys's performance with SOTA works on two practical object detection tasks: cross-domain person detection task (Table~\ref{tab:person}), and cross-domain multi-class detection task (Table~\ref{tab:bdd},~\ref{tab:soda10m}). In each table, we calculate the mAP scores for each local domain alongside the average mAP scores across all domains. Additionally, we report the standard deviation of the local mAP scores to assess the global model's consistency across different domains. 

We have three observations. First, we notice that \sys outperforms baselines in both tasks, and the largest average mAP improvement brought by \sys is 3.03 (on SODA10M).
In addition, compared to other works, \sys enhances the global model's performance the most on the domain with the lowest baseline accuracy or the fewest sample size (the largest mAP improvement on the local domain is 28.23). For example, in the cross-domain person detection task (Table~\ref{tab:person}), the global model achieves an average of 30.57 mAP via FedAvg. However, it only has 13.58 mAP on CityPerson. Moreover, in Table~\ref{tab:soda10m}, the generalized global model achieves much worse mAP on the Rainy domain, 
while \sys improves the performance significantly, 
which greatly benefits the subpopulation client. The reason is that the model exchange enables the model to learn domain-invariant features that are hardly learned in its own domain.
Last but not least,  we observe that \sys achieves small standard deviation scores in Table~\ref{tab:person} and Table~\ref{tab:bdd}. This indicates that, while \sys outperforms all other methods, a low standard deviation across domain mAPs suggests that the final global model is well-generalized and performs equitably across all domains. Conversely, in Table~\ref{tab:soda10m}, \sys demonstrates a larger standard deviation. However, this occurs in a context where mAP scores for all domains show substantial improvement over any of the SOTA methods.

\begin{figure*}[!t]
    \centering
    \subfloat[Multiple Person Datasets]{\includegraphics[width=0.33\linewidth]{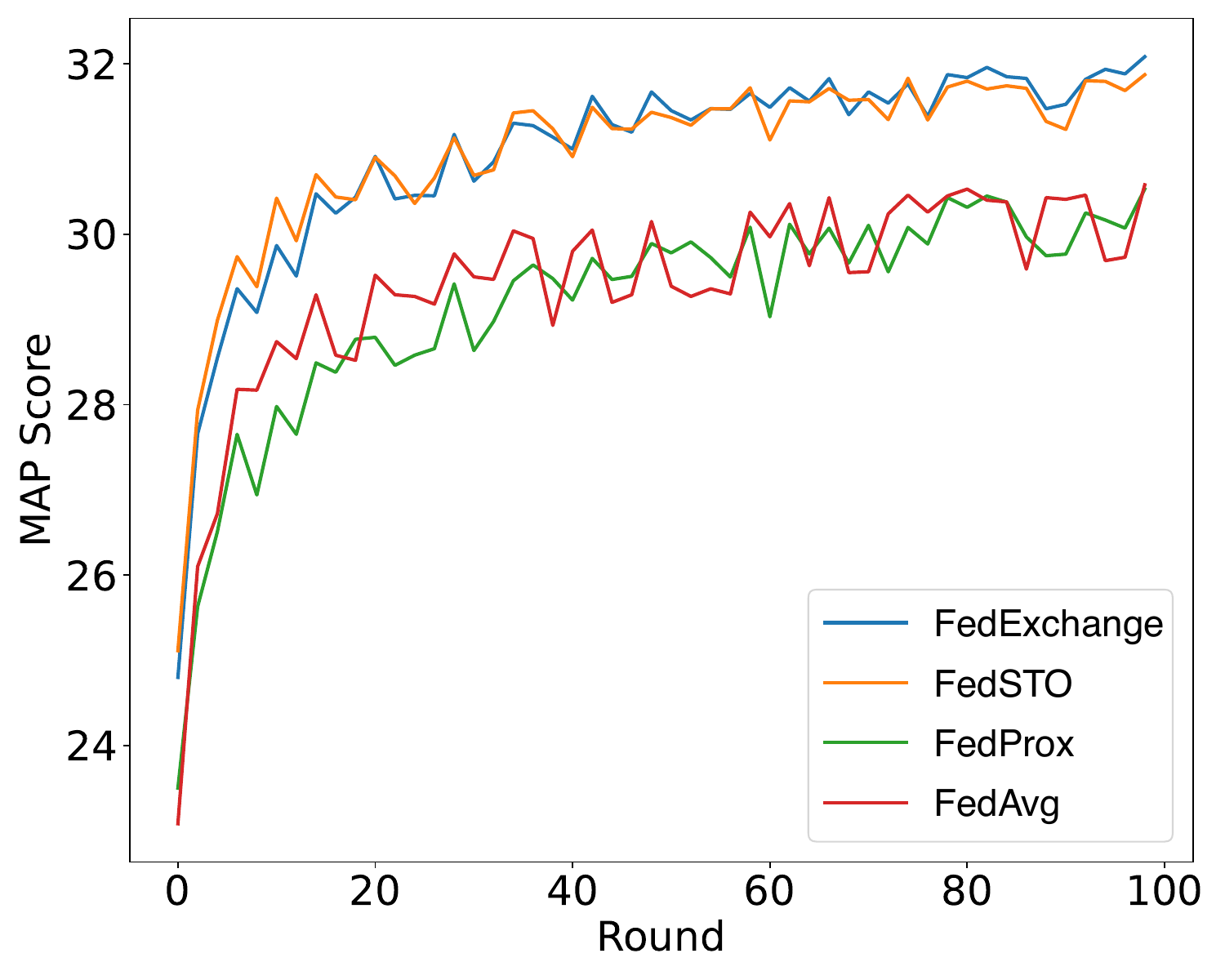}\label{fig:person}}
    \subfloat[BDD100K Dataset]{\includegraphics[width=0.33\linewidth]{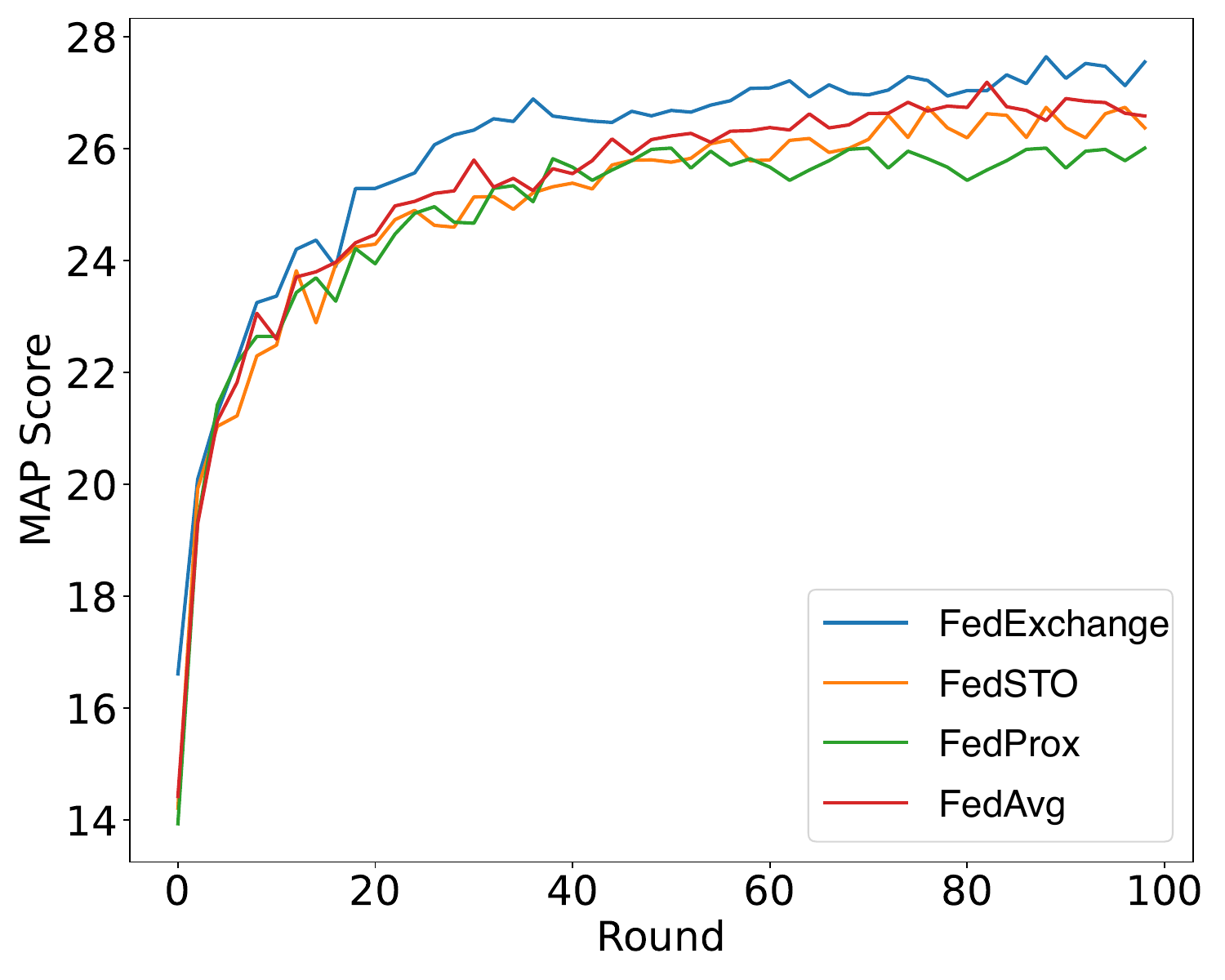}\label{fig:bdd}}
    \subfloat[SODA10M Dataset]{\includegraphics[width=0.33\linewidth]{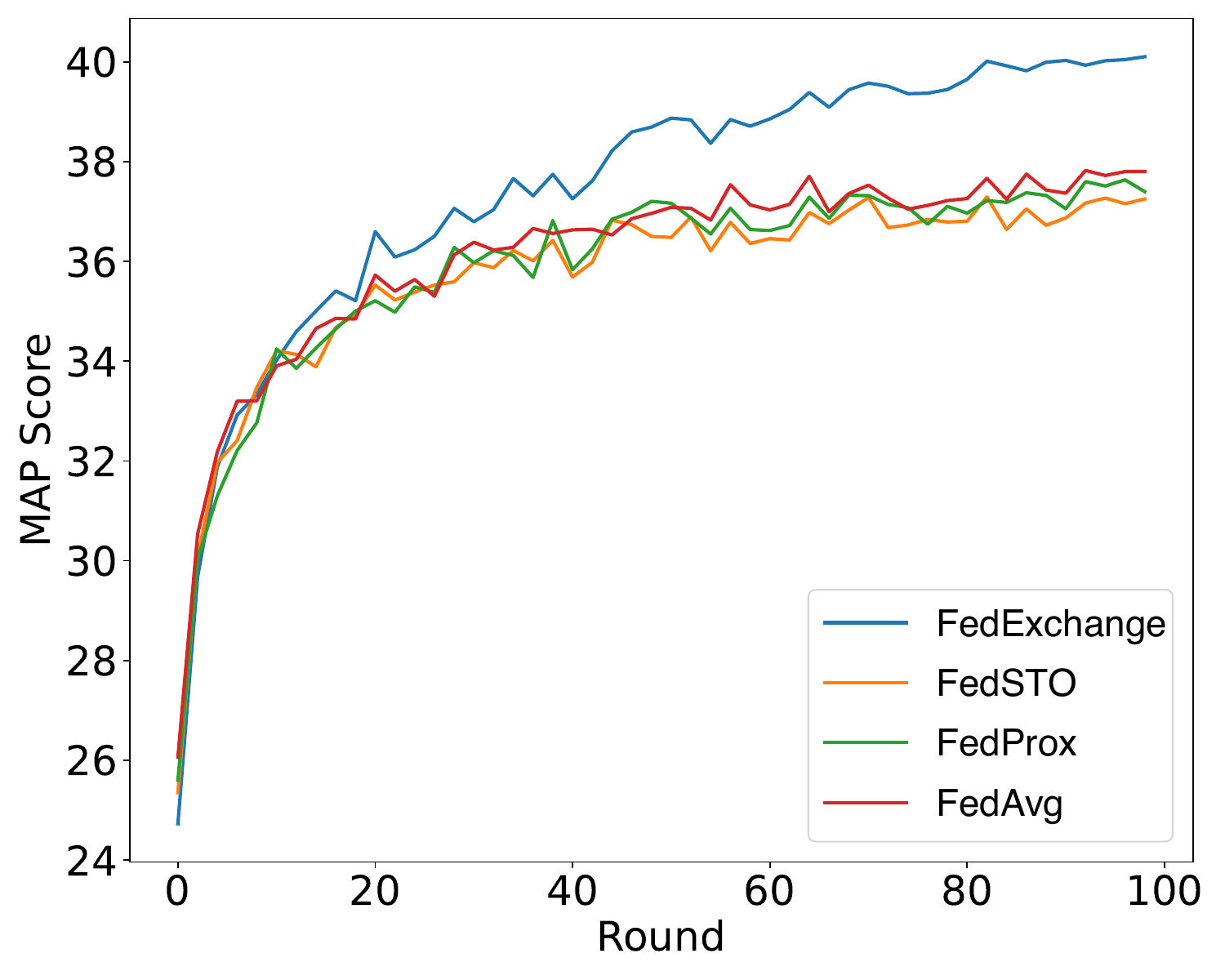}\label{fig:soda}}
    \caption{Evaluation of convergence of \sys vs. baselines. \sys performs better on multiple cross-domain person datasets and achieves better convergence than other methods on BDD100K and SODA100M datasets.}
    \label{fig:convergence}
    \vspace{-0.1in}
\end{figure*}

\begin{table*}[!h]
    \centering
    \setlength{\tabcolsep}{5pt}
    \renewcommand{\arraystretch}{1.2}
    \caption{GPU memory usage \& communication cost per client on BDD100k.}
    \resizebox{0.8\linewidth}{!}{
    \begin{tabular}{c|ccc}
    \toprule  
    Method & Avg. GPU Usage per Client & Local Latency & Global Latency \cr
    
    \midrule
    FedAvg & 4532.3 MB  & 80.5 $\pm$ 62.08 s & 364.43 $\pm$ 43.72 s\cr
    FedProx & 5575.0 MB & 85.5 $\pm$ 66.94 s & 338.24 $\pm$ 11.25 s \cr
    FedSTO & 5315.5  MB & 87.5 $\pm$ 66.15 s  & 373.45$\pm$ 47.17 s\cr
    \midrule
    \sys & 4550.2 MB & 80.1 $\pm$ 62.08 s & 350.53 $\pm$ 42.36 s \cr
    \bottomrule
    \end{tabular}
    }
    \label{tab:overhead}
    \vspace{-0.1in}
\end{table*}

\subsection{Convergence}
\label{sec:convergence}
We show the convergence curve of \sys and other SOTA works on all three evaluations. We show cross-domain person detection in Figure~\ref{fig:person}, and cross-domain multi-class object detection on BDD100k and SODA10M in Figure~\ref{fig:bdd} and~\ref{fig:soda}, respectively. In Figure~\ref{fig:person}, we can observe that from the first evaluation round (which is round 1 in our evaluation), both \sys and FedSTO begin with a higher initial mAP score compared to FedProx and FedAvg. This indicates that \sys has a stronger starting point, suggesting that the proposed model exchange method efficiently makes models learn domain-invariant features from other domains. While FedSTO shows similar initial effectiveness, it does not match \sys’s performance across later rounds. FedProx and FedAvg, in contrast, demonstrate lower mAP scores throughout, showing that they are less effective in handling cross-domain data in FOD.

In both Figure~\ref{fig:bdd} and~\ref{fig:soda}, \sys demonstrates a consistent lead over other methods throughout the rounds. In Figure~\ref{fig:bdd}, \sys shows a rapid increase in mAP score during the early rounds while in~\ref{fig:soda}, it starts with a similar performance to the others but quickly outperforms them, and then maintains a higher accuracy throughout the rounds. Both figures show the ability of \sys to achieve a higher mAP score with a faster and more stable convergence curve. In Figure~\ref{fig:soda}, \sys also shows less fluctuation in later rounds,, which indicates a stable convergence compared to SOTA works.


\begin{table*}[!t]
    \centering
    \setlength{\tabcolsep}{5pt}
    \renewcommand{\arraystretch}{1.2}
    \caption{Ablation study of \sys with $T=2,5,10,50$ on cross-domain person datasets.}
    \resizebox{0.9\linewidth}{!}{
    \begin{tabular}{c|cccc|c}
    \toprule  
     Aggregation Frequency & CityPerson & WiderPerson &  CrowdHuman & COCOPerson & Average\cr
    \midrule
    $T=2$ & \underline{18.70} & \textbf{36.07} & \underline{32.17} & 41.38 & \textbf{32.08}\cr
    $T=5$ & 18.43 & \underline{35.92} & 32.14 & \textbf{41.44} & \underline{31.98} \cr
    $T=10$ & \textbf{19.37} & 33.77 & 31.86  & 40.96 & 31.82\cr
    $T=50$ &  18.73 & 31.47 & \textbf{36.30} & \underline{41.39} & 31.97\cr
    \bottomrule
    \end{tabular}
    }
    \label{tab:T}
    \vspace{-0.1in}
\end{table*}

\begin{table*}[t!]
    \centering
    \scriptsize
    \setlength{\tabcolsep}{5pt}
    \renewcommand{\arraystretch}{1.2}
    \caption{Ablation study of \sys with different dynamic exchange strategies on cross-domain person datasets.}
    \resizebox{0.9\linewidth}{!}{
    \begin{tabular}{c|cccc|c}
    \toprule  
     Exchange Strategy & CityPerson &  WiderPerson & CrowdHuman & COCOPerson  & Average\cr
    \midrule
    Round-robin  & \underline{18.67} & 31.95& \textbf{33.18} & 39.92 & \underline{30.93}\cr
    Random & 17.66& \underline{34.05} & 30.84 & 38.64 & 30.30\cr
    \sys &  \textbf{18.70} & \textbf{36.07} & \underline{32.17} & \textbf{41.38} & \textbf{32.08}\cr
    \bottomrule
    \end{tabular}
    }
    \label{tab:ablation_strategy}
    \vspace{-0.1in}
\end{table*}

\subsection{GPU Memory Usage \& Communication Cost}
\label{sec:GPU}
In this section, we compare the local computational and communication cost of \sys with baseline works. As shown in Table~\ref{tab:overhead}, \sys achieves similar GPU memory usage and local latency to FedAvg. This is because \sys is tailored for practical FOD settings and performs all computationally intensive operations on the server, which 
are more powerful than edge devices. In contrast, both FedProx and FedSTO consume more GPU memory during training,
as they apply additional constraints that require creating and operating tensors of the same size as the model, which also leads to increased latency compared to \sys.

\subsection{Learning in Settings with Limited Training Samples}
\label{sec:scare_training}

\begin{table}
    \centering
    \footnotesize
    \setlength{\tabcolsep}{4pt}
    \renewcommand{\arraystretch}{1.2}
    \caption{\small Evaluation of Limited Training Samples on SODA10M.}
    \begin{tabular}{c|cc}
    \toprule  
    Training set size & \sys & FedAvg \\
    \midrule
    100\% & 40.10 & 37.08 \\
    50\%  & 38.19 & 35.41 \\
    10\%  & 31.35 & 30.18 \\
    \bottomrule
    \end{tabular}
    \label{tab:reduce_sample}
    \vspace{-0.2in}
\end{table}

Since \sys enables models to gain insights from other domains, in this section we aim to evaluate how model exchange impacts model performance when data size on clients are reduced.
We assign each client a training set of varying sizes, reducing the set to 50\% and 10\% of the original dataset size, while fixing the testing set. From Table~\ref{tab:reduce_sample} we see that \sys outperforms FedAvg in all scarce training scenarios. Notably, \sys with a half-sized training sample outperforms FedAvg trained on the full dataset, highlighting the effectiveness of \sys in a scarce-sample training setting. Although each client possesses fewer data, the model exchange employed by \sys enables each model to learn from other domains, mitigating the impact of reduced data. This result demonstrates a potential application of \sys: in situations with limited labeled data, using a smaller training set with \sys can yield a generalized model comparable to one trained on abundant data with FedAvg.

If we further reduce the training set size to only 10\% of the original data, the mAP scores of \sys drop to around 30, similar to the performance of FedAvg with 10\% of the original data. This is likely because the dynamic model exchange in \sys relies on the assumption that model weights reflect the local domain distribution. With only 10\% of the original data, each local domain has too few samples to train an effective model, and the exchanged models may not learn sufficient domain-invariant features from these limited samples. Consequently, this results in a less generalized model across domains.

\subsection{Ablation Study}
\label{sec:ablation}

\subsubsection{Aggregation Frequency}

In this section, we investigate how the exchange and aggregation scheme impact overall FOD utility. Specifically, we set the aggregation frequency 
$T=\{2,5,10,50\}$ and compare the resulting final utility. As shown in Table~\ref{tab:T}, there is no significant decrease in utility when the aggregation frequency increases from 2 to 50 (the largest degradation in mAP is 0.26). This is likely because the model exchanging mechanism introduces some risk of catastrophic forgetting during model exchange. In other words, despite multiple rounds of model exchange, model weights may still largely reflect the latest domain on which the model was trained, which verifies why the mAP scores are similar when increasing the aggregation frequency. Nevertheless, given that \sys achieves its highest performance at $T=2$, users can adjust the aggregation frequency to balance the trade-off between utility and global latency.

\subsubsection{Different Model Exchange Strategies}

In this section, we investigate the relationship between model mAP and different model exchange strategies. Specifically, we consider round-robin and random model exchange as variants alongside the cosine-distance-based metric used in \sys. We evaluate these three schemes on cross-domain person detection tasks, reporting the mAP on each domain along with the overall average mAP in Table~\ref{tab:ablation_strategy}. The round-robin scheme ensures that models are trained on all domains approximately an equal number of times, while the random shuffle scheme provides no such guarantee, as it assigns each model to a domain randomly. From Table~\ref{tab:ablation_strategy}, we observe that compared to random and round-robin model exchange, \sys achieves the highest overall mAP and the best mAP on 3 out of 4 domains. This is because \sys exchanges model orders based on the distance measure, and it ensures that models trained on different domains will be exchanged to learn domain-invariant features, thus leading to a more generalized global model. This result demonstrates the effectiveness of the distance-based exchange scheme, indicating that models benefit from targeted domain-specific insights, thus improving performance across diverse domains.

\section{Conclusion}

In this paper, we introduced \sys, a novel federated object detection (FOD) framework to address cross-domain challenges without incurring additional local computational overhead. By employing a server-side dynamic model exchange strategy, \sys enables local models to benefit from diverse domain data distributions without direct data sharing. The framework alternates between model aggregation and model exchange, using distance-based clustering to facilitate effective cross-domain knowledge transfer.
Extensive evaluations show that \sys significantly improves FOD performance, particularly in challenging domains, while also reducing computational costs relative to baseline methods. We hope this work provides a practical foundation for future research on cross-domain challenges in FOD and inspires further advances in federated learning for complex vision tasks.

\bibliography{aaai2026}

\begin{thebibliography}{50}
\providecommand{\natexlab}[1]{#1}

\bibitem[{Chen et~al.(2019)Chen, Wang, Pang, Cao, Xiong, Li, Sun, Feng, Liu, Xu, Zhang, Cheng, Zhu, Cheng, Zhao, Li, Lu, Zhu, Wu, Dai, Wang, Shi, Ouyang, Loy, and Lin}]{mmdetection}
Chen, K.; Wang, J.; Pang, J.; Cao, Y.; Xiong, Y.; Li, X.; Sun, S.; Feng, W.; Liu, Z.; Xu, J.; Zhang, Z.; Cheng, D.; Zhu, C.; Cheng, T.; Zhao, Q.; Li, B.; Lu, X.; Zhu, R.; Wu, Y.; Dai, J.; Wang, J.; Shi, J.; Ouyang, W.; Loy, C.~C.; and Lin, D. 2019.
\newblock {MMDetection}: Open MMLab Detection Toolbox and Benchmark.
\newblock \emph{arXiv preprint arXiv:1906.07155}.

\bibitem[{Chen et~al.(2023)Chen, Duan, Wang, He, Lu, Dai, and Qiao}]{chen2023vision}
Chen, Z.; Duan, Y.; Wang, W.; He, J.; Lu, T.; Dai, J.; and Qiao, Y. 2023.
\newblock Vision Transformer Adapter for Dense Predictions.
\newblock In \emph{The Eleventh International Conference on Learning Representations}.

\bibitem[{Coates and Ng(2010)}]{ob_robot_2}
Coates, A.; and Ng, A.~Y. 2010.
\newblock Multi-camera object detection for robotics.
\newblock In \emph{2010 IEEE International conference on robotics and automation}, 412--419. IEEE.

\bibitem[{Cordts et~al.(2015)Cordts, Omran, Ramos, Scharw{\"a}chter, Enzweiler, Benenson, Franke, Roth, and Schiele}]{Cityscapes}
Cordts, M.; Omran, M.; Ramos, S.; Scharw{\"a}chter, T.; Enzweiler, M.; Benenson, R.; Franke, U.; Roth, S.; and Schiele, B. 2015.
\newblock The Cityscapes Dataset.
\newblock In \emph{CVPR Workshop on The Future of Datasets in Vision}.

\bibitem[{Deng et~al.(2021)Deng, Li, Chen, and Duan}]{unbiased_KD}
Deng, J.; Li, W.; Chen, Y.; and Duan, L. 2021.
\newblock Unbiased mean teacher for cross-domain object detection.
\newblock In \emph{Proceedings of the IEEE/CVF Conference on Computer Vision and Pattern Recognition}, 4091--4101.

\bibitem[{Deng et~al.(2023)Deng, Xu, Li, and Duan}]{cross_domain_OD_2}
Deng, J.; Xu, D.; Li, W.; and Duan, L. 2023.
\newblock Harmonious Teacher for Cross-Domain Object Detection.
\newblock In \emph{Proceedings of the IEEE/CVF Conference on Computer Vision and Pattern Recognition (CVPR)}, 23829--23838.

\bibitem[{Han et~al.(2021)Han, Liang, Xu, Chen, Hong, Mao, Ye, Zhang, Li, Liang, and Xu}]{soda10m}
Han, J.; Liang, X.; Xu, H.; Chen, K.; Hong, L.; Mao, J.; Ye, C.; Zhang, W.; Li, Z.; Liang, X.; and Xu, C. 2021.
\newblock SODA10M: A Large-Scale 2D Self/Semi-Supervised Object Detection Dataset for Autonomous Driving.
\newblock arXiv:2106.11118.

\bibitem[{He et~al.(2022{\natexlab{a}})He, Wang, Wu, Wang, Li, Li, Gan, Wu, and Qiao}]{cross_domain_OD_1}
He, M.; Wang, Y.; Wu, J.; Wang, Y.; Li, H.; Li, B.; Gan, W.; Wu, W.; and Qiao, Y. 2022{\natexlab{a}}.
\newblock Cross domain object detection by target-perceived dual branch distillation.
\newblock In \emph{Proceedings of the IEEE/CVF Conference on Computer Vision and Pattern Recognition}, 9570--9580.

\bibitem[{He et~al.(2022{\natexlab{b}})He, Wang, Wu, Wang, Li, Li, Gan, Wu, and Qiao}]{dual_branch_KD}
He, M.; Wang, Y.; Wu, J.; Wang, Y.; Li, H.; Li, B.; Gan, W.; Wu, W.; and Qiao, Y. 2022{\natexlab{b}}.
\newblock Cross domain object detection by target-perceived dual branch distillation.
\newblock In \emph{Proceedings of the IEEE/CVF Conference on Computer Vision and Pattern Recognition}, 9570--9580.

\bibitem[{He and Zhang(2020)}]{other_feature_alignment_1}
He, Z.; and Zhang, L. 2020.
\newblock Domain adaptive object detection via asymmetric tri-way faster-rcnn.
\newblock In \emph{Computer Vision--ECCV 2020: 16th European Conference, Glasgow, UK, August 23--28, 2020, Proceedings, Part XXIV 16}, 309--324. Springer.

\bibitem[{Hsu et~al.(2020)Hsu, Tsai, Lin, and Yang}]{center_alignment}
Hsu, C.-C.; Tsai, Y.-H.; Lin, Y.-Y.; and Yang, M.-H. 2020.
\newblock Every pixel matters: Center-aware feature alignment for domain adaptive object detector.
\newblock In \emph{Computer Vision--ECCV 2020: 16th European Conference, Glasgow, UK, August 23--28, 2020, Proceedings, Part IX 16}, 733--748. Springer.

\bibitem[{Hu et~al.(2023)Hu, Yang, Chen, Li, Sima, Zhu, Chai, Du, Lin, Wang et~al.}]{auto}
Hu, Y.; Yang, J.; Chen, L.; Li, K.; Sima, C.; Zhu, X.; Chai, S.; Du, S.; Lin, T.; Wang, W.; et~al. 2023.
\newblock Planning-oriented autonomous driving.
\newblock In \emph{Proceedings of the IEEE/CVF Conference on Computer Vision and Pattern Recognition}, 17853--17862.

\bibitem[{Kim et~al.(2019)Kim, Jeong, Kim, Choi, and Kim}]{other_feature_alignment_2}
Kim, T.; Jeong, M.; Kim, S.; Choi, S.; and Kim, C. 2019.
\newblock Diversify and Match: A Domain Adaptive Representation Learning Paradigm for Object Detection.
\newblock arXiv:1905.05396.

\bibitem[{Kim et~al.(2024)Kim, Lin, Lee, Lau, and Mugunthan}]{FedSTO}
Kim, T.; Lin, E.; Lee, J.; Lau, C.; and Mugunthan, V. 2024.
\newblock Navigating Data Heterogeneity in Federated Learning A Semi-Supervised Federated Object Detection.

\bibitem[{Lepetit and Berger(2000)}]{AR}
Lepetit, V.; and Berger, M.-O. 2000.
\newblock A semi-automatic method for resolving occlusion in augmented reality.
\newblock In \emph{Proceedings IEEE Conference on Computer Vision and Pattern Recognition. CVPR 2000 (Cat. No. PR00662)}, volume~2, 225--230. IEEE.

\bibitem[{Li et~al.(2020{\natexlab{a}})Li, Liu, Su, Xie, Ding, Chen, and Wu}]{cross-domain_fL_3}
Li, S.; Liu, C.~H.; Su, L.; Xie, B.; Ding, Z.; Chen, C. L.~P.; and Wu, D. 2020{\natexlab{a}}.
\newblock Discriminative Transfer Feature and Label Consistency for Cross-Domain Image Classification.
\newblock \emph{IEEE Transactions on Neural Networks and Learning Systems}.

\bibitem[{Li et~al.(2020{\natexlab{b}})Li, Sahu, Talwalkar, and Smith}]{FL_2}
Li, T.; Sahu, A.~K.; Talwalkar, A.; and Smith, V. 2020{\natexlab{b}}.
\newblock Federated learning: Challenges, methods, and future directions.
\newblock \emph{IEEE signal processing magazine}.

\bibitem[{Li et~al.(2020{\natexlab{c}})Li, Sahu, Zaheer, Sanjabi, Talwalkar, and Smith}]{FedProx}
Li, T.; Sahu, A.~K.; Zaheer, M.; Sanjabi, M.; Talwalkar, A.; and Smith, V. 2020{\natexlab{c}}.
\newblock Federated optimization in heterogeneous networks.
\newblock \emph{Proceedings of Machine learning and systems}, 2: 429--450.

\bibitem[{Li et~al.(2022)Li, Dai, Ma, Liu, Chen, Wu, He, Kitani, and Vajda}]{cross_domain_OD_3}
Li, Y.-J.; Dai, X.; Ma, C.-Y.; Liu, Y.-C.; Chen, K.; Wu, B.; He, Z.; Kitani, K.; and Vajda, P. 2022.
\newblock Cross-Domain Adaptive Teacher for Object Detection.
\newblock In \emph{Proceedings of the IEEE/CVF Conference on Computer Vision and Pattern Recognition (CVPR)}, 7581--7590.

\bibitem[{Lin et~al.(2014)Lin, Maire, Belongie, Hays, Perona, Ramanan, Doll{\'a}r, and Zitnick}]{COCO}
Lin, T.-Y.; Maire, M.; Belongie, S.; Hays, J.; Perona, P.; Ramanan, D.; Doll{\'a}r, P.; and Zitnick, C.~L. 2014.
\newblock Microsoft coco: Common objects in context.
\newblock In \emph{Computer Vision--ECCV 2014: 13th European Conference, Zurich, Switzerland, September 6-12, 2014, Proceedings, Part V 13}.

\bibitem[{Liu et~al.(2020)Liu, Long, Zhang, Yu, Dong, and Xiao}]{AR_2}
Liu, D.; Long, C.; Zhang, H.; Yu, H.; Dong, X.; and Xiao, C. 2020.
\newblock ARShadowGAN: Shadow Generative Adversarial Network for Augmented Reality in Single Light Scenes.
\newblock In \emph{2020 IEEE/CVF Conference on Computer Vision and Pattern Recognition (CVPR)}, 8136--8145.

\bibitem[{Liu et~al.(2016)Liu, Anguelov, Erhan, Szegedy, Reed, Fu, and Berg}]{SSD}
Liu, W.; Anguelov, D.; Erhan, D.; Szegedy, C.; Reed, S.; Fu, C.-Y.; and Berg, A.~C. 2016.
\newblock \emph{SSD: Single Shot MultiBox Detector}, 21–37.

\bibitem[{Maiettini et~al.(2020)Maiettini, Pasquale, Rosasco, and Natale}]{ob_robot_1}
Maiettini, E.; Pasquale, G.; Rosasco, L.; and Natale, L. 2020.
\newblock On-line object detection: a robotics challenge.
\newblock \emph{Autonomous Robots}, 44(5): 739--757.

\bibitem[{McMahan et~al.(2023)McMahan, Moore, Ramage, Hampson, and y~Arcas}]{FedAvg}
McMahan, H.~B.; Moore, E.; Ramage, D.; Hampson, S.; and y~Arcas, B.~A. 2023.
\newblock Communication-Efficient Learning of Deep Networks from Decentralized Data.
\newblock arXiv:1602.05629.

\bibitem[{Nguyen et~al.(2022)Nguyen, Pham, Pathirana, Ding, Seneviratne, Lin, Dobre, and Hwang}]{health_2}
Nguyen, D.~C.; Pham, Q.-V.; Pathirana, P.~N.; Ding, M.; Seneviratne, A.; Lin, Z.; Dobre, O.; and Hwang, W.-J. 2022.
\newblock Federated learning for smart healthcare: A survey.
\newblock \emph{ACM Computing Surveys (Csur)}, 55(3): 1--37.

\bibitem[{Oh et~al.(2011)Oh, Hoogs, Perera, Cuntoor, Chen, Lee, Mukherjee, Aggarwal, Lee, Davis et~al.}]{surveillance}
Oh, S.; Hoogs, A.; Perera, A.; Cuntoor, N.; Chen, C.-C.; Lee, J.~T.; Mukherjee, S.; Aggarwal, J.; Lee, H.; Davis, L.; et~al. 2011.
\newblock A large-scale benchmark dataset for event recognition in surveillance video.
\newblock In \emph{CVPR 2011}, 3153--3160. IEEE.

\bibitem[{Oquab et~al.(2023)Oquab, Darcet, Moutakanni, Vo, Szafraniec, Khalidov, Fernandez, Haziza, Massa, El-Nouby et~al.}]{oquab2023dinov2}
Oquab, M.; Darcet, T.; Moutakanni, T.; Vo, H.; Szafraniec, M.; Khalidov, V.; Fernandez, P.; Haziza, D.; Massa, F.; El-Nouby, A.; et~al. 2023.
\newblock Dinov2: Learning robust visual features without supervision.
\newblock \emph{arXiv preprint arXiv:2304.07193}.

\bibitem[{Ren et~al.(2016)Ren, He, Girshick, and Sun}]{ren2016faster}
Ren, S.; He, K.; Girshick, R.; and Sun, J. 2016.
\newblock Faster R-CNN: Towards real-time object detection with region proposal networks.
\newblock \emph{IEEE transactions on pattern analysis and machine intelligence}, 39(6): 1137--1149.

\bibitem[{Rezaeianaran et~al.(2021)Rezaeianaran, Shetty, Aljundi, Reino, Zhang, and Schiele}]{other_feature_alignment_3}
Rezaeianaran, F.; Shetty, R.; Aljundi, R.; Reino, D.~O.; Zhang, S.; and Schiele, B. 2021.
\newblock Seeking similarities over differences: Similarity-based domain alignment for adaptive object detection.
\newblock In \emph{Proceedings of the IEEE/CVF International Conference on Computer Vision}, 9204--9213.

\bibitem[{Saito et~al.(2019)Saito, Ushiku, Harada, and Saenko}]{strong_weak_alignment}
Saito, K.; Ushiku, Y.; Harada, T.; and Saenko, K. 2019.
\newblock Strong-weak distribution alignment for adaptive object detection.
\newblock In \emph{Proceedings of the IEEE/CVF conference on computer vision and pattern recognition}, 6956--6965.

\bibitem[{Sandler et~al.(2019)Sandler, Howard, Zhu, Zhmoginov, and Chen}]{mobilenet_v2}
Sandler, M.; Howard, A.; Zhu, M.; Zhmoginov, A.; and Chen, L.-C. 2019.
\newblock MobileNetV2: Inverted Residuals and Linear Bottlenecks.
\newblock arXiv:1801.04381.

\bibitem[{Shao et~al.(2018)Shao, Zhao, Li, Xiao, Yu, Zhang, and Sun}]{Crowdhuman}
Shao, S.; Zhao, Z.; Li, B.; Xiao, T.; Yu, G.; Zhang, X.; and Sun, J. 2018.
\newblock CrowdHuman: A Benchmark for Detecting Human in a Crowd.
\newblock \emph{arXiv preprint arXiv:1805.00123}.

\bibitem[{Shokri and Shmatikov(2015)}]{FedSGD}
Shokri, R.; and Shmatikov, V. 2015.
\newblock Privacy-preserving deep learning.
\newblock In \emph{Proceedings of the 22nd ACM SIGSAC conference on computer and communications security}.

\bibitem[{Sun et~al.(2021)Sun, Huo, Yang, and Bai}]{cross-domain_fL}
Sun, B.; Huo, H.; Yang, Y.; and Bai, B. 2021.
\newblock Partialfed: Cross-domain personalized federated learning via partial initialization.
\newblock \emph{Advances in Neural Information Processing Systems}, 34: 23309--23320.

\bibitem[{Swaminathan, Silver, and Akilan(2024)}]{jetson_nano}
Swaminathan, T.~P.; Silver, C.; and Akilan, T. 2024.
\newblock Benchmarking Deep Learning Models on NVIDIA Jetson Nano for Real-Time Systems: An Empirical Investigation.
\newblock arXiv:2406.17749.

\bibitem[{Wang et~al.(2024)Wang, Bian, Zhang, Chen, and Xu}]{cross-domain_fL_2}
Wang, L.; Bian, J.; Zhang, L.; Chen, C.; and Xu, J. 2024.
\newblock Taming Cross-Domain Representation Variance in Federated Prototype Learning with Heterogeneous Data Domains.
\newblock arXiv:2403.09048.

\bibitem[{Xie et~al.(2019)Xie, Yu, Wang, Wang, and Zhang}]{multi-level_alignment}
Xie, R.; Yu, F.; Wang, J.; Wang, Y.; and Zhang, L. 2019.
\newblock Multi-level domain adaptive learning for cross-domain detection.
\newblock In \emph{Proceedings of the IEEE/CVF international conference on computer vision workshops}, 0--0.

\bibitem[{Xu et~al.(2021)Xu, Glicksberg, Su, Walker, Bian, and Wang}]{health_1}
Xu, J.; Glicksberg, B.~S.; Su, C.; Walker, P.; Bian, J.; and Wang, F. 2021.
\newblock Federated learning for healthcare informatics.
\newblock \emph{Journal of healthcare informatics research}, 5: 1--19.

\bibitem[{Yang et~al.(2023)Yang, Hui, Yuan, Gong, and Cao}]{FL_classification_1}
Yang, Y.; Hui, B.; Yuan, H.; Gong, N.; and Cao, Y. 2023.
\newblock $\{$PrivateFL$\}$: Accurate, differentially private federated learning via personalized data transformation.
\newblock In \emph{32nd USENIX Security Symposium (USENIX Security 23)}, 1595--1612.

\bibitem[{Yang et~al.(2024)Yang, Lee, Dariush, Cao, and Lo}]{yang2024anomalyruler}
Yang, Y.; Lee, K.; Dariush, B.; Cao, Y.; and Lo, S.-Y. 2024.
\newblock Follow the Rules: Reasoning for Video Anomaly Detection with Large Language Models.
\newblock In \emph{Proceedings of the European Conference on Computer Vision (ECCV)}.

\bibitem[{Yu et~al.(2020)Yu, Chen, Wang, Xian, Chen, Liu, Madhavan, and Darrell}]{BDD100K}
Yu, F.; Chen, H.; Wang, X.; Xian, W.; Chen, Y.; Liu, F.; Madhavan, V.; and Darrell, T. 2020.
\newblock Bdd100k: A diverse driving dataset for heterogeneous multitask learning.
\newblock In \emph{Proceedings of the IEEE/CVF conference on computer vision and pattern recognition}.

\bibitem[{Yuan et~al.(2022)Yuan, Hui, Yang, Burlina, Gong, and Cao}]{FL_classification_2}
Yuan, H.; Hui, B.; Yang, Y.; Burlina, P.; Gong, N.~Z.; and Cao, Y. 2022.
\newblock Addressing heterogeneity in federated learning via distributional transformation.
\newblock In \emph{European Conference on Computer Vision}, 179--195. Springer.

\bibitem[{Yuan et~al.(2024)Yuan, Paul, Aucott, Burlina, and Cao}]{FL_classification_3}
Yuan, H.; Paul, W.; Aucott, J.; Burlina, P.; and Cao, Y. 2024.
\newblock PFedEdit: Personalized Federated Learning via Automated Model Editing.

\bibitem[{Zhang et~al.(2019)Zhang, Xie, Wan, Xia, Li, and Guo}]{WiderPerson}
Zhang, S.; Xie, Y.; Wan, J.; Xia, H.; Li, S.~Z.; and Guo, G. 2019.
\newblock WiderPerson: A Diverse Dataset for Dense Pedestrian Detection in the Wild.
\newblock \emph{IEEE Transactions on Multimedia (TMM)}.

\bibitem[{Zhang et~al.(2022)Zhang, Cao, Jia, and Gong}]{finance_2}
Zhang, Z.; Cao, X.; Jia, J.; and Gong, N.~Z. 2022.
\newblock Fldetector: Defending federated learning against model poisoning attacks via detecting malicious clients.
\newblock In \emph{Proceedings of the 28th ACM SIGKDD Conference on Knowledge Discovery and Data Mining}, 2545--2555.

\bibitem[{Zheng et~al.(2021)Zheng, Yan, Gou, and Wang}]{finance_1}
Zheng, W.; Yan, L.; Gou, C.; and Wang, F.-Y. 2021.
\newblock Federated meta-learning for fraudulent credit card detection.
\newblock In \emph{Proceedings of the Twenty-Ninth International Conference on International Joint Conferences on Artificial Intelligence}, 4654--4660.

\bibitem[{Zheng et~al.(2020)Zheng, Huang, Liu, and Wang}]{cross_domain_OD_4}
Zheng, Y.; Huang, D.; Liu, S.; and Wang, Y. 2020.
\newblock Cross-domain Object Detection through Coarse-to-Fine Feature Adaptation.
\newblock In \emph{2020 IEEE/CVF Conference on Computer Vision and Pattern Recognition (CVPR)}, 13763--13772.

\bibitem[{Zhu et~al.(2019)Zhu, Pang, Yang, Shi, and Lin}]{other_feature_alignment_4}
Zhu, X.; Pang, J.; Yang, C.; Shi, J.; and Lin, D. 2019.
\newblock Adapting Object Detectors via Selective Cross-Domain Alignment.
\newblock In \emph{Proceedings of the IEEE Conference on Computer Vision and Pattern Recognition}, 687--696.

\bibitem[{Zhuang, Chen, and Lyu(2023)}]{zhuang2023foundation}
Zhuang, W.; Chen, C.; and Lyu, L. 2023.
\newblock When foundation model meets federated learning: Motivations, challenges, and future directions.
\newblock \emph{arXiv preprint arXiv:2306.15546}.

\bibitem[{Zhuang et~al.()Zhuang, Xu, Chen, Li, and Lyu}]{COALA}
Zhuang, W.; Xu, J.; Chen, C.; Li, J.; and Lyu, L. ????
\newblock COALA: A Practical and Vision-Centric Federated Learning Platform.
\newblock In \emph{Forty-first International Conference on Machine Learning}.

\end{thebibliography}
\newpage
\appendix
\section*{Appendix}
\label{sec:appendix}

\renewcommand{\algorithmicrequire}{\textbf{Input:}}
\renewcommand{\algorithmicensure}{\textbf{Output:}}
    \begin{algorithm}[!h] \scriptsize
        \caption{\sys}
        \label{algorithm}
        \begin{algorithmic}[1]
        \Require pre-trained global backbone $f$, client $i$'s initialized global decoder $g_l^0$ at round 0, communication round $R$,  source domain training set $D_{s}$, aggregation frequency $T$
        \Ensure Optimized decoder $g^r_l$ for $r$-th round
        \For{each client $k \in N$ \text{ in parallel}}
        \State $f_l^i \gets f$ 
        \State $g^0_l \gets WARM\_UP(D_s, f_l^i, g^0_l)$ 
        \EndFor
        \For{$r$ in range(1,$R+1$)}
        \For{each client $i \in C$ \text{ in parallel}}
        \State $g_{l,i}^r \gets Local\_training(f_l^i, g_{l,i}^{r-1})$
                \State $Upload\_model(g_{l,i}^r)$
        \EndFor
        \If{ $r\% T == 0$} 
            \State $g^r_g \gets global\_aggregate(G)$
        \Else
            \State $C_1, C_2 \gets Server\_Clustering(G)$
            \State $G\_list \gets Online\_Exchange(C_1,C_2)$
        \EndIf 
        \For{each client $i \in N$ \text{ in parallel}}
        \If{G\_list is not None}
        \State $g^r_{l,i} \gets G\_list[i]$
        \Else
        \State $g^r_{l,i} \gets g^r_g$
        \EndIf
        \EndFor
        \EndFor
        \end{algorithmic}
    \end{algorithm}

\subsection*{Experimental Setup}

We conduct our experiments on NVIDIA RTX-6000 GPUs using PyTorch 2.3.0 and Python 3.10.13. We build a simulation framework based on COALA~(\citealt{COALA}) and mmdetection~(\citealt{mmdetection})   for FL object detection evaluation.

\paragraph*{Datasets.}
For datasets, we include four person detection datasets and two driving datasets. For cross-domain person detection evaluation, we use all four person datasets, assigning each client one dataset as their local domain. Additionally, for cross-domain multi-class object detection evaluation, we evaluate each driving dataset separately, and clients are assigned with portions of data samples based on either scene or weather. Datasets details are introduced in Appendix. All images are resized to 480$\times$640 and the batch size is set to be 16 for both BDD100k and SODA10M. For person datasets, images are resized to 384$\times$512 and the batch size is set to be 64. 

\begin{icompact}
    \item \textbf{Person Detection Datasets.} i). CityPerson~(\citealt{Cityscapes}). A subset of Cityscapes for person detection. It contains 2975 images for training and 500 images for validation. ii). WiderPerson~(\citealt{WiderPerson}). A benchmark dataset contains dense pedestrians with various occlusions. All image are selected from a wide range of scenarios and the dataset contains 8,000 training samples and 1,000 validation samples. iii). CrowdHuman~(\citealt{Crowdhuman}). A large person detection dataset with high diversity. It contains 15,000 training images and 4,370 validation images. iv). COCOPerson. A subset split from MSCOCO~(\citealt{COCO}) as the person 
    detection dataset. It contains training images and validation images. We take 6,000 training images as training set.
    \item \textbf{BDD100k}~(\citealt{BDD100K}). A multi-tasks driving dataset that contains 70,000 training images and 10,000 validation images. The dataset includes five classes:car, pedestrian, bus, truck, and traffic sign, and spans across six scenarios: city street, gas station, tunnel, parking lot, highway, and residential. We categorize gas station, tunnel, and parking lot as \textit{mixed} scene due to their limited sample sizes, while assigning the remaining samples to clients based on their respective scenes.
    \item \textbf{SODA10M}~(\citealt{soda10m}). SODA10M is a large 2D autonomous driving dataset for object detection. It contains six classes, namely, pedestrian, cyclist, car, truck, tram, and tricycle. We split the labeled dataset based on weather condition, i.e., clear, overcast, and rainy, and each client is assigned one weather condition.  
\end{icompact}

\paragraph{Baselines and Metric.} Due to the fact that there lack of state-of-the-art (SOTA) literature on cross-domain FOD, we compare \sys with three works that are highly related. Specifically, we compare \sys with FedSTO~(\citealt{FedSTO}), which is the latest semi-supervised FOD work. Moreover, we also compare \sys with FedAvg~(\citealt{FedAvg}), which is the standard federated learning approach, and FedProx~(\citealt{FedProx}), which is an FL framework designed to improve model utility under data heterogeneity. Note that given our focus on a practical foundation model setting, we do not compare against FL approaches that rely on specific architectural assumptions, such as FedBN. 
We conduct evaluations under a semi-supervised setting following FedSTO. As for the performance metric, we use the mean average precision (mAP) as the model utility indicator in our evaluation.

\paragraph{Model Structure and Hyper-Parameters.} We implement the foundation model for object detection with a Dinov2-pretrained ViT backbone~(\citealt{oquab2023dinov2}) adapted for object detection using ViT-adapter~(\citealt{chen2023vision}). We pre-train the adapter part on the COCO dataset while keeping Dinov2-ViT frozen to preserve its generalization ability. The decoder consists of a Feature Pyramid Network (FPN) as the neck, and a Faster R-CNN head~(\citealt{ren2016faster}) consists of a Region Proposal Network (RPN) head and a Region of Interest (ROI) head. We use the AdamW optimizer with a learning rate of 1e-3. The training process includes a warm-up phase of 5 rounds, and the total number of training rounds is set to 100. For evaluation, we use $T=2$ as our default setting.

\section*{Performance compared to local models}

\begin{table*}[!t]
    \centering
    \setlength{\tabcolsep}{5pt}
    \renewcommand{\arraystretch}{1.2}
    \caption{Performance of \sys Compared to Local Models on Cross-Domain Person Datasets}
    \resizebox{0.8\linewidth}{!}{
    \begin{tabular}{c|cccc|c}
    \toprule  
     \multirow{2}{*}{\diagbox{Model}{Dataset}} & COCOPerson & CityPerson &  CrowdHuman & WiderPerson & \multirow{2}{*}{Average}\cr
     \cline{2-5}
     & Domain 1 & Domain 2 & Domain 3  & Domain 4\cr
     
    \midrule
     Domain 1 & 45.34	&18.01	&30.47	& 21.92 & 28.93\cr
    Domain 2 & 21.98	&23.33	&24.01	&26.19	&23.88 \cr
    Domain 3 & 39.96	&16.61	&38.06	&26.2	&30.21\cr
    Domain 4 &  22.86	&18.69	&23.01	&42.96	&26.88\cr
    \hline
    Strongest & 45.34	&23.33	&38.06	&42.96	&37.42\cr
    \sys & 41.38	&18.7	&32.17	&36.07	&32.08\cr
    \bottomrule
    \end{tabular}
    }
    \label{tab:supp_1}
\end{table*}

\begin{table*}[!t]
    \centering
    \setlength{\tabcolsep}{5pt}
    \renewcommand{\arraystretch}{1.2}
    \caption{Performance of \sys Compared to Local Models on BDD100K Datasets}
    \resizebox{0.8\linewidth}{!}{
    \begin{tabular}{c|cccc|c}
    \toprule  
     \multirow{2}{*}{\diagbox{Model}{Dataset}} & CityStreet & Mixed &  Highway & Residential & \multirow{2}{*}{Average}\cr
     \cline{2-5}
     & Domain 1 & Domain 2 & Domain 3  & Domain 4\cr
     
    \midrule
     Domain 1 & 25.53	&20.06	&20.57	&25.2	&22.84\cr
    Domain 2 & 18.17	&19.24	&14.72	&16.45	&17.15 \cr
    Domain 3 & 24.75	&21.07	&23.45	&23.7	&23.24\cr
    Domain 4 &  26.25	&24.01	&22.19	&27.41	&24.97\cr
    \hline
    Strongest & 26.25	&24.01	&23.45	&27.41	&25.28\cr
    \sys & 28.75	&29.14	&26.83	&25.85	&27.64\cr
    \bottomrule
    \end{tabular}
    }
    \label{tab:supp_2}
\end{table*}

\begin{table}[!t]
    \centering
    \small
    \setlength{\tabcolsep}{5pt}
    \renewcommand{\arraystretch}{1.2}
    \caption{Performance of \sys Compared to Local Models on SODA10M Datasets}
    \begin{tabular}{c|ccc|c}
    \toprule  
     \multirow{2}{*}{\diagbox{Model}{Dataset}} & Clear & Overcast &  Rainy & \multirow{2}{*}{Average}\cr
     \cline{2-4}
     & Domain 1 & Domain 2 & Domain 3  \cr
     
    \midrule
     Domain 1 & 28.44	&29.52	&45.26 & 34.41\cr
    Domain 2 &  27.01	&48.75	&41.43 & 39.06\cr
    Domain 3 & 20.23	&22.46	&68.31	&37.00\cr
    \hline
    Strongest & 28.44	&48.75	&68.31	&48.50\cr
    \sys & 30.86	&38.16	&75.08	&48.03\cr
    \bottomrule
    \end{tabular}
    \label{tab:supp_3}
\end{table}

Our goal is to train a universal model that achieves the best average performance across all domains. In Table~\ref{tab:supp_1}, for the person detection task, \sys outperforms any locally trained model by up to 8.21\%, highlighting its capability to learn domain-invariant features. For object detection tasks, we compare \sys with the best-performing model in each domain. In Table~\ref{tab:supp_2}, \sys achieves the best performance in 3 out of 4 domains, and in Table~\ref{tab:supp_3}, it outperforms others in 2 out of 3 domains, with average improvements of up to 8.04\% and 23.4\%, respectively.

These results demonstrate that \sys, by leveraging knowledge from multiple domains, can surpass models trained solely on local data. Notably, even for tasks with abundant training samples, models may still struggle to learn features that generalize across domains. In contrast, domain-invariant features transferred from other domains enhance the model’s ability to recognize diverse scenarios without compromising performance, leading to significantly improved cross-domain utility.

\end{document}